\documentclass{article}
\usepackage{authblk}

\usepackage[a4paper, total={14.5cm, 24cm}]{geometry}
\usepackage[utf8]{inputenc} 
\usepackage[T1]{fontenc}    
\usepackage{hyperref}
\usepackage{graphicx}
\usepackage{url} 
\usepackage{booktabs}       
\usepackage{amsfonts}       
\usepackage{amsmath}
\usepackage{amsthm}
\usepackage{siunitx} 
\usepackage{changes}

\usepackage{xcolor}
\usepackage{todonotes}
\usepackage{listings}

\usepackage{changes}
\usepackage{changes}
\definechangesauthor[name={Gilles}, color=blue]{g}

\usepackage{natbib}
\newcommand{\msf}[1]{\text{\textsf{\small{#1}}}}
\newcommand{\tf}[1]{\textsf{\small{#1}}}
\newcommand{\wdt}[1]{\textsf{\small{\textcolor{gray}{#1}}}}

\title{Understanding Wikidata Qualifiers: An Analysis and Taxonomy.   
}
\author[1,2]{ Gilles Falquet\thanks{\texttt{gilles.falquet@unige.ch}} }
  \author[1]{Sahar Aljalbout\thanks{\texttt{saharaljalbout@gmail.com}} }

\affil[1]{University of Geneva, Switzerland }
\affil[2]{ISRI, Geneva, Switzerland}

\date{}

\begin{document}

\maketitle

\subsection*{Abstract}
This paper presents an in-depth analysis of Wikidata qualifiers, focusing on their semantics and actual usage, with the aim of developing a taxonomy that addresses the challenges of selecting appropriate qualifiers, querying the graph, and making logical inferences. The study evaluates qualifier importance based on frequency and diversity, using a modified Shannon entropy index to account for the "long tail" phenomenon. By analyzing a Wikidata dump, the top 300 qualifiers were selected and categorized into a refined taxonomy that includes contextual, epistemic/uncertainty, structural, and additional qualifiers. The taxonomy aims to guide contributors in creating and querying statements, improve qualifier recommendation systems, and enhance knowledge graph design methodologies. The results show that the taxonomy effectively covers the most important qualifiers and provides a structured approach to understanding and utilizing qualifiers in Wikidata.

\section{Introduction}
A Wikidata graph is a set of \textit{statements} that describe \textit{entities}. An entity can be either an \textit{item} or a \textit{property} or a \textit{datatype}. A statement asserts that a given entity (the subject) possesses a particular property with a specific value. Statements can be qualified with additional property-value pairs and it should contain at least one reference to an information source. In this paper we use expressions of the form
$$(s, p, v)[q_1: v_1, \dots , q_n: v_n]$$
to denote a statement in which $s$ is the subject, $p$ and $v$ form the main property-value pair, and the $q_i: v_i$'s are the qualifying property-value pairs. 

In \cite{wikibase2025} the pair $(p,v)$ is called the main snak of the statement and the $q_i: v_i$ pairs are called the qualifier snaks (or qualifiers for short). In this paper, slightly departing from this terminology we will call $p$ the \textit{property} of the statement and $q_1, ..., q_n$ the \textit{qualifiers} of the statement.  Thus, a qualifier is defined as a property used in a $q_i: v_i$ qualifying pair (qualifier snak).

To make the presentation more readable, items and properties will be noted by their English label rather than their identifier (e.g.  \tf{David Bowman} instead of \tf{Q4886422}). Accordingly, the statement 
\begin{quote}
    George C. Scott was married to Colleen Dewhurst from 1960 until their divorce in 1965
\end{quote}
will be denoted as:
\begin{equation}
\label{eq:scott}
\begin{array}{l}
 (\msf{\wdt{George C. Scott}, spouse, \wdt{Colleen Dewhurst}}) \\
\quad [\msf{start time}: \wdt{1960},  \msf{end time}: \wdt{1965},  \msf{end cause}: \wdt{divorce}] \\ 
\end{array}
\end{equation}
instead of
\begin{equation}
\label{eq:scott-spouse}
\begin{array}{l}
 (\msf{\wdt{Q182450}, P26, \wdt{Q253916}}) \\
\quad [\msf{P580}: \wdt{1960},  \msf{P582}: \wdt{1965},  \msf{P1534}: \wdt{Q93190}] \\ 
\end{array}
\end{equation}

The Wikidata statements form a multi-qualified\footnote{also known as multi-attributed knowledge graph} knowledge graph because each statement may have several values for each qualifier. For example in the statement
\begin{equation}
\label{eq:lorentz}
\begin{array}{l}
(\msf{\wdt{Pare Lorentz}, notable work, \wdt{The Plow That Broke the Plains}}) \\
\quad [\msf{publication date :} \wdt{10 May 1936}, \msf{subject has role :} \wdt{ film director}, \\
\quad \msf{subject has role :} \wdt{ screenwriter},  \msf{subject has role :} \wdt{ film editor}] \\
\end{array}
\end{equation}
the \tf{subjet has role} qualifier has three distinct values.

The introduction of qualifiers makes the Wikidata model more compact than a simple graph model such as RDF 1.1 \cite{cyganiak2014}. In a simple graph model qualifying a statement, e.g. to add a temporal validity interval, requires and explicit reification of the statement or another design pattern, such as Singleton Properties \cite{nguyen2014don}
,  NdFluents \cite{gimenez2017ndfluents}, Hoganification \cite{hogan2018context}, etc., 
that in one way or another adds complexity to the model.

Based on its usage by Wikidata contributors, the concept of qualifier clearly fulfills a real need within the community. As of the first of January 2025,   \num{2240} properties were used as qualifiers, appearing in approximately 20\% of the statements (\num{331631397} statements out of  \num{1624567397} are qualified)
\footnote{\texttt{script: src/gen-stats/stmt\_count.py}. This notation is used throughout the article to refer to the scripts used to produce the results presented in the article. They can be found on Github at \url{https://github.com/cui-ke/wikidata-qualifiers} together with their output.}\footnote{These figures do not include the ``example'' statements, as explained in Section \ref{sect:qualif-freq-div}}. However, this abundance presents a spectrum of challenges: 
\begin{itemize}
    \item Selecting the appropriate qualifiers during the creation of new statements becomes a overwhelming task for the contributors. 
    \item When querying the graph, it's imperative to consider specific qualifiers to formulate queries that yield meaningful results. This becomes particularly crucial with qualifiers that restrict the validity of a statement to some spatial, temporal, or organizational domain, or with qualifiers that express uncertainty about the statement.
    \item Establishing logical inferences requires elucidating how qualifier values transition from conditions to conclusions.  This problem has been deeply studied in  \cite{aljalbout2023handling}
\end{itemize}

We propose to address the aforementioned issues, by defining a taxonomy of qualifiers based on their actual usage type in the Wikidata graph (shown in Figure \ref{fig:taxonomy}).  

\paragraph{Methodology  and structure of the article:} 
To establish a taxonomy of qualifiers, a thorough examination of their semantics and actual usage is essential. Given the extensive number of qualifiers this study focuses on the most important ones. In section 2, we show that a important qualifier is characterized by a high frequency and/or a high diversity of use. We describe how we adapted a diversity index based on Shannon entropy to obtain a diversity index that is suitable for Wikidata qualifiers. To create the qualifier taxonomy 
we first analyzed a complete Wikidata dump to select the 300 most important qualifiers (in terms of frequency and diversity). Then, starting from an existing taxonomy, introduced in \cite{Patel-Schneider2018} , we analyzed the selected qualifiers, checked if they fit into one of the taxonomy's classes, and updated the taxonomy  when necessary. 
The qualifier analysis was based on their descriptions and their actual use in Wikidata\footnote{For this studiy we used the 2025-01-01 Wikidata json dump. All the analysis results can be found on Github at \url{https://github.com/cui-ke/wikidata-qualifiers/tree/main/stats/2025-01-01}}, (the properties of the statements they qualify and their values). This process yielded a taxonomy comprising qualifier categories and subcategories, the characteristics of which are presented in Section\ref{sect:results}. 
In order to characterise this taxonomy, we demonstrate how qualifiers are distributed across the categories and examine cases where apparent classification confusion occurs(Section \ref{sect:taxo-charact}). In Section \ref{sect:use}, we  explain how this taxonomy can be used to select the right qualifiers when creating new statements and to formulate correct queries over Wikidata. In conclusion, we also show possible applications of this taxonomy beyond those mentioned in the problem statement, in particular in the design of new knowledge graphs.

\section{Qualifiers, Frequency and Diversity\label{sect:qualif-freq-div}}

In the introduction we defined a qualifier as a property that is used to qualify statements. In the present study we must restrict definition for two reasons:
\begin{enumerate}
    \item In principle, every property $p$ should be accompanied by an example, which is a statement of the form 
    $$(p, \msf{Wikidata property example}\footnotemark, \mathit{subject})[p: value]$$
    \footnotetext{In addition to \tf{Wikidata property example} (P1855), example statements may be build with \tf{Wikidata property example for senses} (P5977), \tf{Wikidata property example for media} (P6685), \tf{Wikidata property example for properties} (P2271), \tf{Wikidata property example for lexemes}(P5192), \tf{Wikidata property example for forms} (P5193)}
    that represents the statement 
    $$(subject, p, value).$$
    For example 
    $$(\msf{highest point}, \msf{Wikidata property example}, \msf{Norway})[\msf{highest point}: \msf{Galdhøpiggen}] $$
    indicates that the statement
    $$(\msf{Norway}, \msf{highest point},  \msf{Galdhøpiggen})$$
    exemplifies the use of the \msf{highest point} property.
    
    Therefore every property appears at least once as a qualifier in an example statement. But the example statements cannot be considered as ordinary statements. They are conceptually meta-statements (statements about statements). This is why we won't consider them when analyzing the qualifiers. 
    \item Some properties may have scope constraints that prevent them from being used as qualifiers. For instance, the statement 
    $$(\msf{instance of}, \msf{property constraint}, \msf{property scope constraint})[\msf{property scope}: \msf{as main value}]$$
    indicates that the \msf{instance of} (P31) property can only appear as predicate in a statement and not as qualifier.
    
    Therefore, even though these properties may actually used to qualify statements, this is considered misuse and they should not be counted as qualifiers 
    \footnote{In the 2025-01-01 dump we found that 880  properties used as qualifiers violated a ``property scope'' constraint (compare \path{stats/2025-01-01/q-freq.json} and \path{stats/2025-01-01/allowed-as-qualifiers.csv})}. 
\end{enumerate}
Taking these two points into account we consider that a qualifier is a property that 1) qualifies at least one statement that is not an exemple statement and 2) is not disallowed by a property scope constraint. Applying this definition on the 2025-01-01 Wikidata dump we found \num{1357} properties that are qualifiers and \num{881} properties used as qualifiers in non-example statements that violate a scope constraint.

To evaluate the importance of a qualifier, we consider both the number of statements it qualifies (its frequency) and the diversity of the different statements it qualifies. 

\subsection{Frequency}

We define the frequency $F(q)$ of a qualifier $q$ as the number of statements in which $q$ appears at least once. 

\begin{figure}[ht]
    \centering
    \includegraphics[width=0.98\textwidth]{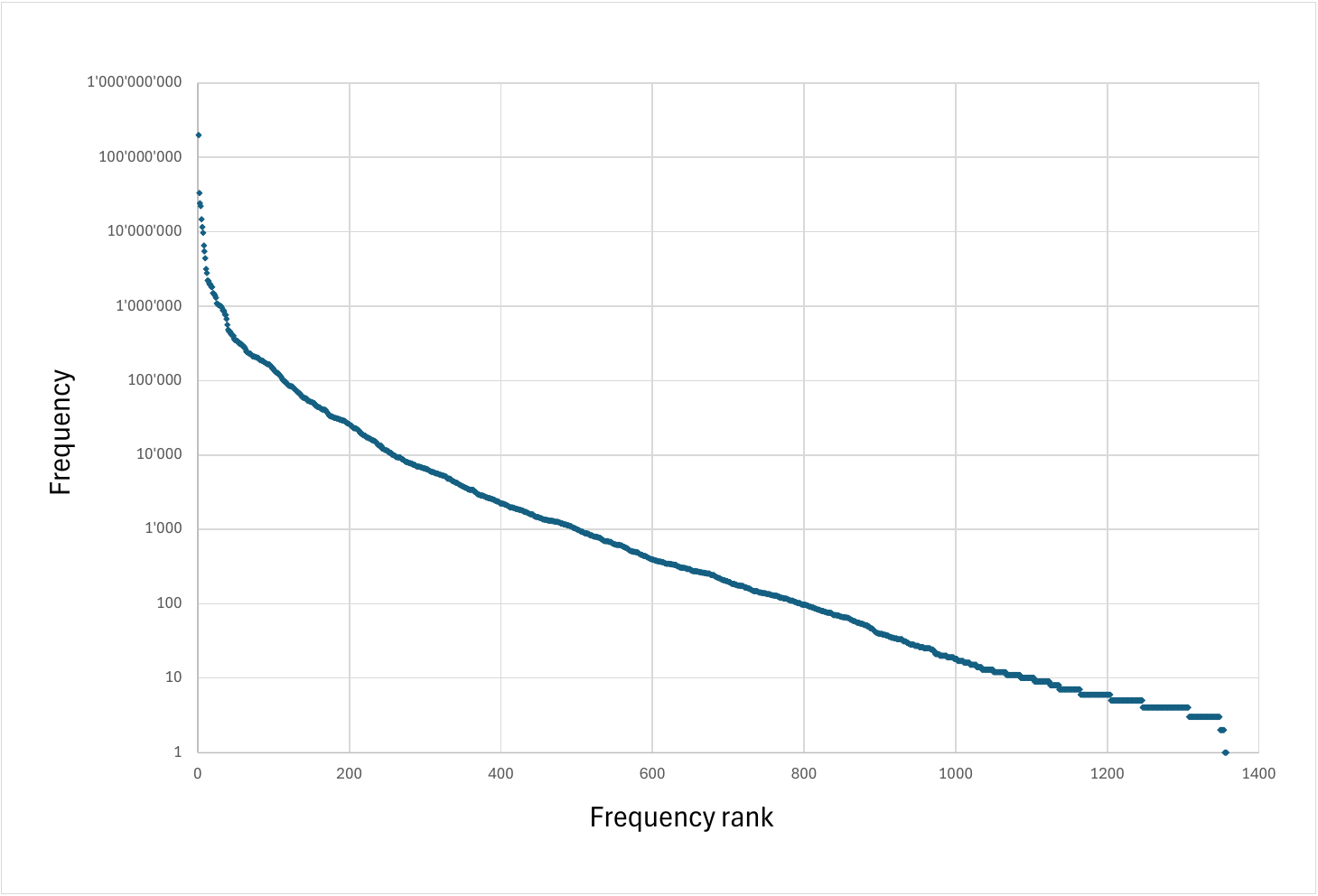}
    \caption{Frequency as a function of the frequency rank for the Wikidata qualifiers}
    \label{fig:freqs}
\end{figure}

We performed a frequency analysis on a full Wikidata dump to get an idea of the frequency distribution of the qualifiers.
\footnote{script: \path{src/import/extract_p_q_freq.py}, output: \path{stats/2025-01-01/p-freq.json}} 
Figure \ref{fig:freqs} plots the qualifier frequencies in order of decreasing value.   It shows that, as a function of its position in the ranking, the frequency of a qualifier decreases slightly more quickly than a negative exponential. In fact, 41 qualifiers (3.02\%) have a frequency higher than \num{1000000} whereas 579 qualifiers (42.7\%) have a frequency lower that \num{100}.
%

\subsection{Diversity}
The importance of a qualifier is not necessarily reflected by this frequency alone. A qualifier such as \tf{astronomical\ filter} (P1227), for example, is extremely frequent (\num{33145290} statements qualified), but it is used almost exclusively to qualify statements about the \tf{apparent\ magnitude} (P1215) of stars (\num{33145248} statements). Conversely, the qualifier \tf{valid\ in\ place} (P3005) is much less frequent, yet it is used to qualify statements with properties such as  \tf{host} (P2975), \tf{number of cases} (P1603), \tf{number\ of\ recoveries} (P8010), \tf{instance of} (P31), \tf{box office} (P2142), \tf{number of deaths} (P1120), and 339 others. Therefore the importance of a qualifier involves two dimensions: the number of qualified statements (frequency) and the number of distinct  properties appearing  in these statements (diversity). 

\begin{figure}[ht]
    \centering
    \includegraphics[width=0.98\textwidth]{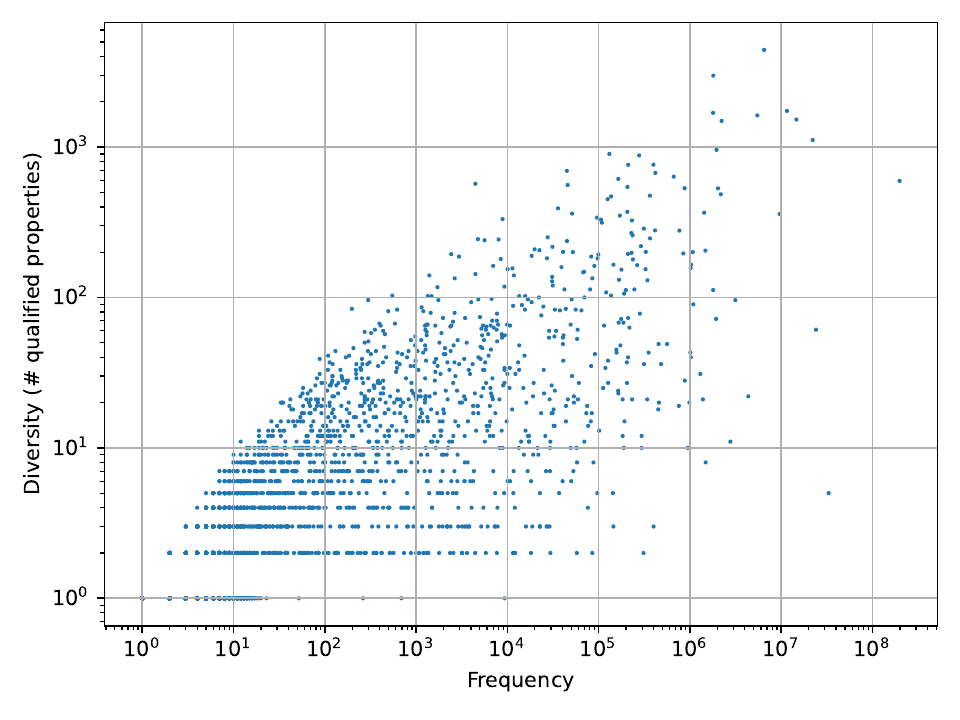}
    \caption{Frequency/diversity plot of the Wikidata qualifiers where the diversity is defined as the number of distinct properties qualified by a qualifier}
    \label{fig:freq-nbqprop}
\end{figure}

However, the number of distinct properties in the qualified statements (qualified properties) is not directly usable as a diversity indicator. To illustrate this, consider the \tf{catalog (P972)} qualifier. This qualifier is applied to 61 different properties across \num{24052733} statements. However, of these 61 properties, 37 appear fewer than ten times each, meaning that fewer than ten statements are qualified by \tf{catalog} for each of these 37 properties. 

Since this ``long tail'' phenomenon occurs with most of the Wikidata qualifiers it is advisable to use a well-known diversity index such as the Hill numbers (or effective number of species)\footnote{\url{https://en.wikipedia.org/wiki/Diversity_index}}, which is commonly used to measure biodiversity and that takes into account the relative frequencies (abundances) of the species. 

For this study, we have chosen to use the Hill numbers of order 1. If a dataset contains $R$ types (e.g. species) with relative frequencies $p_1, ..., p_R$, the Hill number of order 1 is  $$^1\!D:=\exp\left(-\sum_{i = 1}^{R} p_i \ln(p_i)\right).$$ 
This index is the exponential of the Shannon index or entropy.

 If the $R$ types in a dataset are equally abundant, with relative frequencies of $1/R$, then the dataset's diversity is $R$. In other words, a dataset whose diversity index is $d$ has the same diversity as a dataset comprising $d$ equally abundant types.

To apply this formula to a qualifier $q$ let us first define
\begin{itemize}
    \item the frequency $F(p, q)$ of $p$ qualified by $q$, as the number of statements with property $p$ that are qualified by $q$,
    \item  the properties qualified by $q$,  $P(q)$, as the set of properties $p$ such that $F(p, q) > 0$ (at least one statement with property $p$ is qualified by $q$).
    \item the relative frequency $pr(p,q)$ of $p$ for $q$ as
 $$pr(p,q) = F(p, q)/\sum_{p' \in P(q)}F(p', q).$$
 
\end{itemize}

Then the diversity index of a qualifier $q$  is defined as
 $$D(q)=\exp\left(-\sum_{p \in P(q)} pr(p,q) \ln(pr(p,q))\right).$$
The diversity index behaves in the same way as the entropy: If $P(q)$ contains $n$ properties, the index is maximal when the properties have equal relative frequencies of $1/n$. In this case, the index is equal to $n$. The index is minimal when one property has a relative frequency close to 1, while the others are close to 0; in this case, the index is close to 1.  As mentioned above, a diversity index of $d$ means that the qualifier has a diversity that is approximately equal to that of a qualifier which qualifies $d$ properties with the same relative frequencies.

 For example, consider the qualifier \tf{genomic assembly} (\tf{Q659}), whose property frequencies are shown in Table \ref{tab:abq659},, its Shannon index $D(\msf{Q659})$ is 3.26. This effectively reflects the fact that it is essentially used with the properties \tf{genomic end},  \tf{genomic start},  \tf{strand orientation}, and  \tf{chromosome}. Its use with the other four properties is accidental or incorrect.
\begin{table}[ht]
    \centering
    \begin{tabular}{|c|l|r|l|}
    \hline
    $p$ & Property name &  $F(p, \msf{Q656})$ & $pr(p, \msf{Q656})$  \\
    \hline
    P645 & genomic end & \num{514956}& 0.3457874454	\\
    P644 &  genomic start & \num{514955}& 0.3457867739 	\\
    P2548 & strand orientation & \num{422711} & 0.2838459147	\\
    P1057 & chromosome & \num{36553} & 0.0245449484	\\
    P2043 &  length & 46 & 0.0000308885 	\\
    P1855  & Wikidata property example & 3 & 0.0000020144 	\\
    P3331 & HGVS nomenclatur & 2 & 0.0000013429 	\\
    P4844 & research intervention & 1 & 0.0000006714  \\
    \hline
    \end{tabular}
\caption{Frequencies and relative frequencies of the properties in P(\msf{Q656}) (\tf{genomic assembly})}
    \label{tab:abq659}
\end{table} 

Nevertheless, this index is still not entirely satisfactory, as some properties tend to overshadow the others, drastically reducing the diversity index. For instance, the \tf{series ordinal} qualifier has the frequencies shown in Table  \ref{tab:f1545}. Due to the very high frequency of \tf{author name string}, the diversity index is only 2.03. This does not reflect the fact that this qualifier is very frequently used with many other properties. Notably, it qualifies 99\% of statements with the property \tf{expressed in} and 94\% of statements with the property \tf{author'}.
\begin{table}[ht]
    \centering
    \begin{tabular}{|c|l|r|}
        \hline
        $p$ &  Property name &$F(p, \msf{P1545})$  \\
        \hline
P2093 & author name string & \num{148015200} \\
P50 & author & \num{31769610} \\
P2860 & cites work & \num{14782837} \\
P5572 & expressed in & \num{1056827} \\
P179 & part of the series & \num{745165} \\
P735 & given name & \num{739798} \\
P527 & has part(s) & \num{352459} \\
P793 & significant event & \num{210584} \\
P4908 & season & \num{170291} \\
P5753 & ideographic description sequence & \num{90169} \\
P361 & part of & \num{83594} \\
P26 & spouse & \num{79612} \\
P734 & family name & \num{68516} \\
P658 & tracklist & \num{43553} \\
P710 & participant & \num{37559} \\
        etc & & \\
        \hline
    \end{tabular}
    \caption{Frequencies of the properties qualified by \tf{series ordinal (P1545)}}
    \label{tab:f1545}
\end{table}

Thus we opted to replace the frequency $F(p,q)$ of a property $p$ for a qualifier $q$ by the proportional frequency $$PF(p, q) = F(p, q)/GF(p)$$
where $GF(p)$, the global frequency of $p$, is the number of statements having property $p$. 
$PF(p,q)$ is the proportion of the statements having property $p$ that are qualified by $q$. 

\begin{table}[ht]
    \centering
    \begin{tabular}{|c|l|r|r|r|}
        \hline 
        $p$ &  Property name & $GF(p)$ &$F(p, \msf{P1545})$ & $PF(p, \msf{P1545})$\\
        \hline
P2093 & author name string & \num{148810279} & \num{148015200} & 0.995\\
P50 & author & \num{33552582} & \num{31769610} & 0.947\\
P2860 & cites work & \num{303125156} & \num{14782837} & 0.049\\
P5572 & expressed in & \num{1058470} & \num{1056827} & 0.998\\
P179 & part of the series & \num{1003261} & \num{745165} & 0.743\\
P735 & given name & \num{7901359} & \num{739798} & 0.094\\
P527 & has part(s) & \num{2405547} & \num{352459} & 0.147\\
P793 & significant event & \num{1129126} & \num{210584} & 0.187\\
P4908 & season & \num{172152} & \num{170291} & 0.989\\
P5753 & ideographic description sequence & \num{90169} & \num{90169} & 1.000\\
P361 & part of & \num{5104617} & \num{83594} & 0.016\\
P26 & spouse & \num{830942} & \num{79612} & 0.096\\
P734 & family name & \num{5346143} & \num{68516} & 0.013\\
P658 & tracklist & \num{53797} & \num{43553} & 0.810\\
P710 & participant & \num{860972} & \num{37559} & 0.044\\
        etc.   & ...                 & ...          & ...   & ... \\
        
        \hline
    \end{tabular}
    \caption{Global frequency (GF), frequency (F), and proportional frequencies (PF) of the properties qualified by \tf{series ordinal (P1545)}}
    \label{tab:propf}
\end{table}

For the above example, the proportional frequencies are shown in Table  \ref{tab:propf}.  Using the proportional frequencies instead of the frequencies to compute the diversity index yields a value of 57.64, which better reflects the usage of this qualifier. Interestingly, for \tf{genomic assembly}, the new index remains at 3.41, close to the original value. This is desirable because the last four properties are incorrectly or accidentally qualified.

\begin{figure}[ht]
    \centering
    \includegraphics[width=0.75\linewidth]{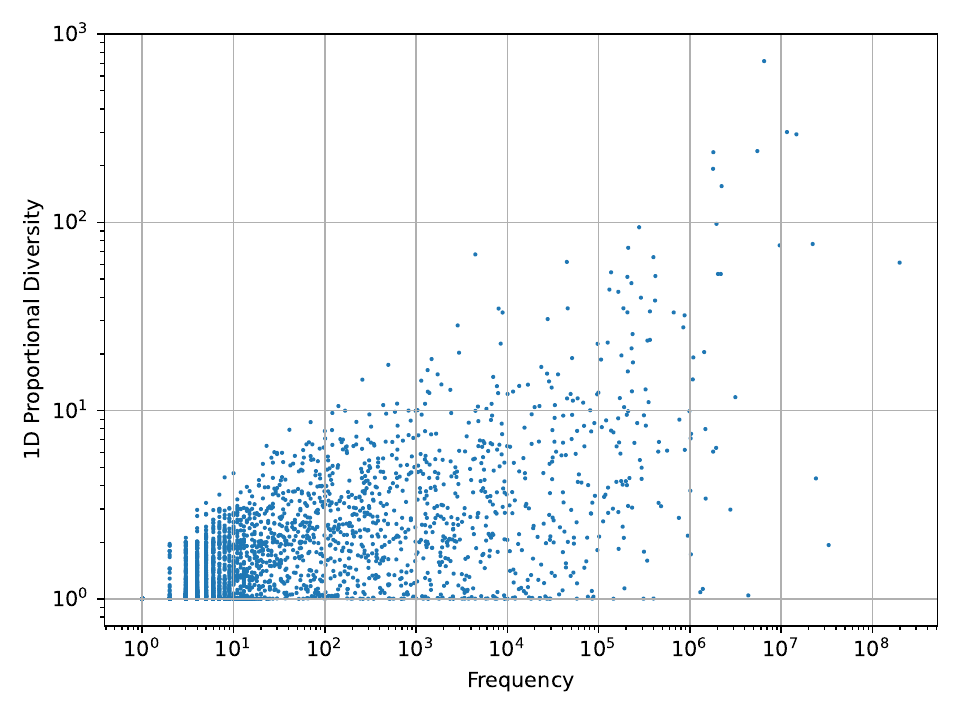}
    \caption{Frequency and $^{1}\!D$ proportional diversity of the Wikidata qualifiers}
    \label{fig:freq-1dp-diversity}
\end{figure}

Figure \ref{fig:freq-1dp-diversity} presents the position of each qualifier based on its frequency and $^{1}\!D$ proportional diversity. It graphically shows that there is no correlation between these two variables.

\section{Categorization Approach for Qualifiers}

The taxonomy construction is based on an analysis of the most important qualifiers, in terms of frequency and diversity, from the  the Wikidata graph. Frequency and diversity computations were carried out by extracting three dictionaries\footnote{script: \path{src/import/extract_p_q_freq}, output: \path{stats/2025-01-01/pfreq.json}, \path{stats/2025-01-01/q-freq.json}. \path{stats/2025-01-01/p-q-freq.json}}  from the 2025-01-01 Wikidata dump:
\begin{description}
    \item[P $\to$ F] associates every property with its frequency (the number of statements with this property)
    \item[Q $\to$  F] associates every qualifier with its frequency (the number of times it appears in a statement) 
    \item[Q $\to$ (P $\to$ F)] associates with each qualifier $q$ a dictionary that associates with each property $p$  the value of $F(p, q)$ (the number of times the qualifier $q$ appears in a statement with property $p$)  
\end{description}

From these data we computed an importance score for each qualifier \footnote{script: \path{src/import/diversity_index.py}, output: \path{stats/2025-01-01/q-diversity-score.csv}}, which is defined as $$\mathit{score}=\mathit{frequency}\times\mathit{diversity}$$


Since the frequency and diversity values are expressed in different units (frequency is a number of statements, diversity is a number of properties) and since their value ranges differ greatly, it is  preferable to combine them by multiplication rather than by weighted addition. This is particularly important to obtain a ranking that is independent of the normalization functions , as explained in, for example, \cite{tofallis2014}. 

We selected the top 300 qualifiers as a basis to  build  a taxonomy of qualifiers. These qualifiers cover  99.6\% of the \num{96281446} qualifications present in the graph, i.e. 99.6\% of the qualifier-value pairs have a qualifier belonging to the selected set.

Each selected qualifier has been analyzed based on its textual description and on the properties it most frequently qualifies. As the taxonomy aims to reflect the actual use of qualifiers, qualifier analysis could not be based solely on the qualifier descriptions. This is because qualifiers are properties that are used in a qualifying role. However, the description of many properties refers only to their role as statement properties; their role as qualifiers is not described. In fact, only 148 of the 300 selected qualifiers have a description containing a descriptive string, usage instructions, or examples that explicitly mention their role as a qualifier. Furthermore, the textual descriptions are generally limited to one sentence and do not specify the intended use of the qualifiers precisely.

To determine how qualifiers are actually used, we consulted the $\mathbf{Q}\to (\mathbf{P} \to \mathbf{F})$  dictionary to identify the most frequently qualified properties. This generally provides  a good indication of the actual (operational) meaning of the qualifier for the Wikidata contributors. 
For instance, the properties most frequently qualified by \tf{language of work or name} (P407) are (in descreasing ordre of frequency) \tf{official website} (P856),	
\tf{described at URL} (P973),		
\tf{work available at URL} (P953),		
\tf{Fandom article ID} (P6262),		
\tf{pronunciation audio} (P443),		
\tf{has edition or translation} (P747),		
\tf{YouTube video ID} (P1651),		
\tf{voice actor} (P725),		
\tf{curriculum vitae URL} (P8214), etc. Almost all of them have values that are URLs or other external identifiers or other resources that are not Wikidata items \footnote{By the way, \tf{has edition or translation} has a value which is a Wikidata item that should be described by a  \tf{language of work or name} property. Therefore the qualification by \tf{language of work or name} is modeling mistake (redundancy)}. This means that the \tf{language of work or name} qualifier does not describe  the statement itself  but the external resource designated by the statement's value. We will see in the next section that this kind of qualifiers fall into the \tf{External Entity Description} category.

If this was insufficient to determine the meaning of the qualifier, we queried the Wikidata graph to find the values most frequently associated with it. This was necessary for qualifiers such as \tf{sourcing circumstances} that exhibit high diversity. In this case the list of qualified properties is very heterogeneous and does not help in finding the meaning of the qualifier.

The categorisation of qualifiers started with Patel-Schneider's classification  \cite{patelcontextualization}  that distinguishes between \emph{contextual} and \emph{additional} qualifiers. Contextual qualifiers restrict the validity of a claim made in a statement to a particular context. For example: 
\begin{equation}
\label{eq:ibm}
\begin{array}{l}
(\wdt{IBM}, \msf{official website}, \wdt{https://www.ibm.com/de-de/})\\
\quad \quad [\msf{applies to jurisdiction}: \wdt{Germany}]
\end{array}
\end{equation}
means that \emph{\url{https://www.ibm.com/de-de/} is the official website of IBM} only in German jurisdiction.
Additional qualifiers, on the other hand, have no influence on the semantics of the statement, they provide supplementary information, as in .
\begin{equation}
\label{eq:dandelion}
\begin{array}{l}
(\wdt{common dandelion}, \msf{taxon name}, \wdt{Taraxacum officinale})\\
\quad \quad [\msf{taxon author}: \wdt{Friedrich Heinrich Wiggers}]
\end{array}
\end{equation}

We also considered other ways to characterize statements, such as  causality, uncertainty, confidence, provenance, and context description. These have been amply studied in the knowledge representation literature. For an introduction to some of these aspects, see, for instance, \cite{porter2008} .

\section{Results\label{sect:results}}

\begin{figure}[ht]
    \centering
    \includegraphics[width=0.75\linewidth]{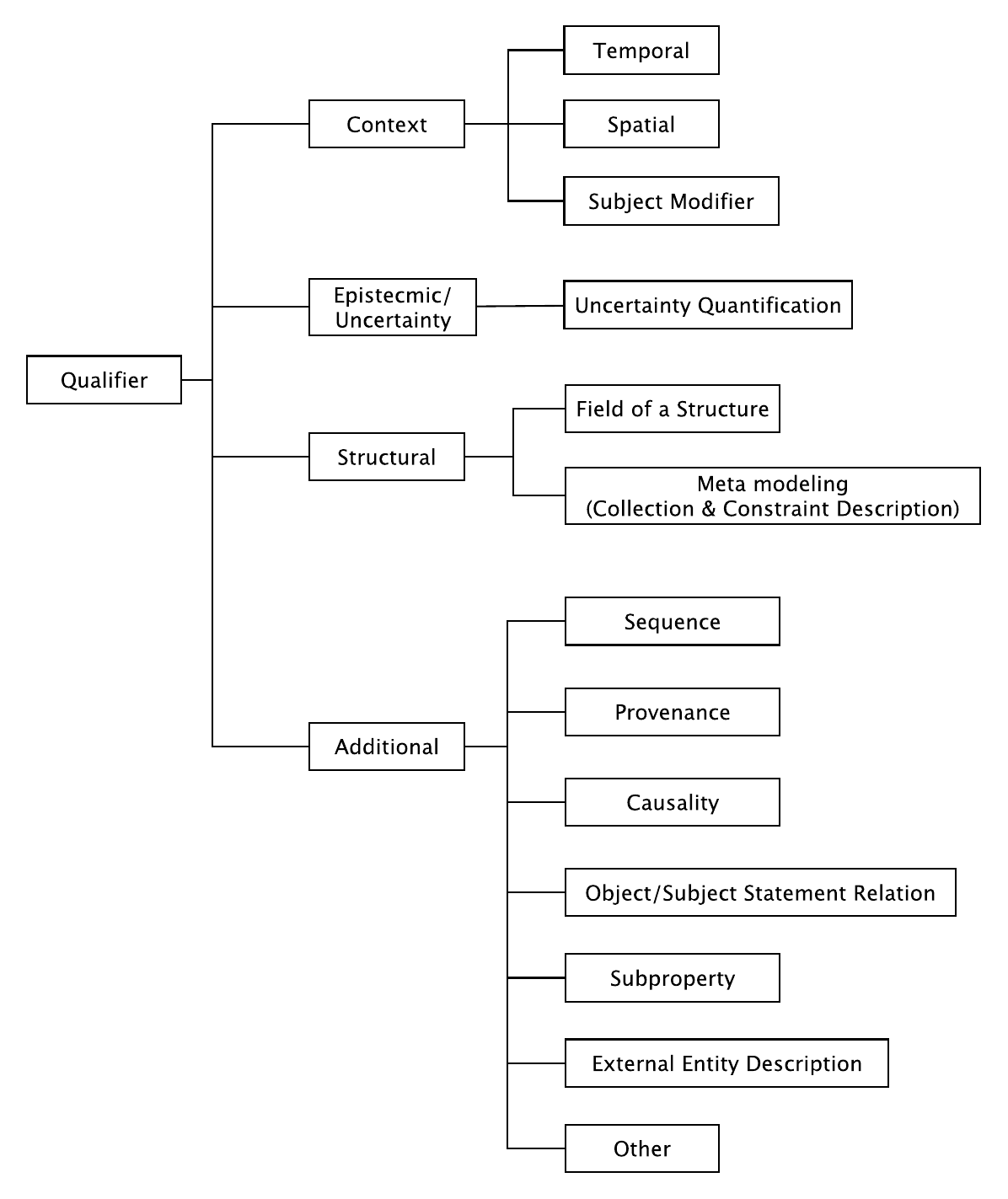}
    \caption{Qualifier taxonomy}
    \label{fig:taxonomy}
\end{figure}

The classification we propose is a refinement and an extension of the classification by Patel-Schneider \cite{Patel-Schneider2018}. By studying the most important qualifiers, as explained in the previous section, we found that some non-contextual qualifiers are not additional. These qualifiers cannot be removed without alterning the meaning of the statement. These qualifiers either carry epistemic information,  generally defining the uncertainty (or imprecision) of a statement), or participate in specifying a structured value or defining  a data structure (lists, categories) or constraint. We also found that the contextual and additional qualifier category  can be refined into several non-overlapping subcategories. This resulted in the taxonomy shown in Figure \ref{fig:taxonomy}.

In the rest of this section we describe the characteristics of each qualifier category.

 \subsection{Context Qualifiers (aka Validity)} 
    
These qualifiers limit the validity of a statement to a specified context. Without this restriction, the statement would be universally valid for the given subject and object. 
    
A context can be defined along several independent dimensions, such as time, space, or an abstract region (e.g. a jurisdiction or an organization), In each dimension, the validity qualifiers define a subset of all possible values, whether temporal, spatial, etc.  Therefore, the validity of a statement is the Cartesian product of the validity subsets for all the considered dimensions.

 \subsubsection*{Temporal context}
        
 A temporal context is usually defined using  qualifiers such as \tf{point in time}, \tf{start time}, \tf{end time}, or \tf{valid in period}. This is generally a time interval, which can be reduced to an instant, as in 
        $$
            \label{eq:peoria}
            \begin{array}{l}
            (\textsf{\wdt{Peoria}, population, \wdt{190985}}) \\
            \quad [\textsf{point in time}: \wdt{1 April 2020},  ...] \\ 
            \end{array}
        $$
The absence of a temporal context specification can have various meanings. For example, if the statement is about something that cannot change over time, such as the location of the 1969 Woodstock music festival, the notion of temporal validity does not apply. In other words, the statement is always valid. If the statement's subject is an endurant item (something whose properties can change over time), the absence of a temporal context specification usually means that the statement is valid for the subject's entire lifetime up to the present day., e.g.        $$
        \begin{array}{l}
        (\msf{\wdt{Eiffel tower}, number of floor, \wdt{3}}).
        \end{array}
        $$
However, it may also mean that the statement is currently valid, but was not always valid, or may not always be valid, e.g. $$
        \begin{array}{l}
        (\msf{\wdt{Aletsch Glacier}, thickness, \wdt{900 metre}})
        \end{array}
        $$
        
\subsubsection*{Spatial context}
        
Qualifiers such as \tf{country} or \tf{valid in place} define the spatial validity of a statement. See for example statement \ref{eq:ibm}.

\subsubsection*{Subject modifiers}

These qualifiers specify the validity of a statement by either extending or restricting the subject to which it applies. Typical qualifiers in this category are: \tf{applies to part}, \tf{including},  and \tf{excluding}. 
%

        $$
            \label{eq:dance-of-time}
            \begin{array}{l}
            (\textsf{\wdt{Dance of Time}, creator, \wdt{Jean-Baptiste Lepaute}}) \\
            \quad [\textsf{applies to part: \wdt{clock}},  ...] \\ 
            \end{array}
        $$


\paragraph*{Remark.}
The qualifier description can sometimes be misleading. For example,  the description of \tf{catalog} contains ``as a qualifier of P528 -- catalog for which the 'catalog code' is valid''. This suggest that it is a context qualifier, as it defines the validity domain of a code. However, \tf{catalog} can hardly be considered a context qualifier because, unlike other context qualifiers, it has a low diversity (4.36) , and mostly qualifies \tf{catalog code} (P528) and only a few other identification number properties. It is much more natural to consider that \tf{catalog} is a field of a structured value consisting of a pair (\tf{catalog code}, \tf{catalog}).  

\subsection{Epistemic/Uncertainty} 

A few qualifiers, in particular \textsf{sourcing circumstances (P1480)} and \textsf{ nature of statement (P5102) } relate to our knowledge of a statement or the extent to which it has been validated. They can indicate that the statement's value is imprecise, as in: 
    $$
    \begin{array}{l}
    (\msf{\wdt{Rome}, inception, \wdt{21 April 753 BCE}})[\msf{sourcing circumstances: \wdt{circa}}],
    \end{array}
    $$
    or uncertain, as in: 
    $$
    \begin{array}{l}
         (\msf{\wdt{Cyrillic script}, creator, \wdt{Constantine of Preslav}})  \\
         \quad [\msf{sourcing circumstances: \wdt{hypothesis} }].
    \end{array}
    $$
    These qualifiers can also express a level of confidence in the statement, with values such as \tf{official}, \tf{unofficial}, \tf{disputed}, \tf{cannot be confirmed by other sources}, or even the falsity of a statement

Some of these qualifiers are intended to quantify uncertainty in a value.
    $$
    \begin{array}{l}
    (\msf{\wdt{Plato}, date of birth, \wdt{420s BCE}}) \\
    \quad [\msf{earliest date: \wdt{428 BCE}, latest date: \wdt{427 BCE}}]
    \end{array}
    $$
The most frequently used among these qualifiers are \tf{latest date}, \tf{earliest date}, \tf{latest start date}, \tf{earliest end date}, \tf{latest end date}.  Note that these qualifiers do not define a temporal validity context, instead, they indicate that the true value of a \tf{date}, \tf{start date}, or \tf{end date} lies somewhere within the given interval.

\subsection{Structural}

Structural qualifiers participate in the definition of data structures, at either the data type level (metamodeling) or at the value level (a field within a structured value). 

\subsubsection*{Metamodeling}
    
These qualifiers are used to enrich the Wikidata model by defining categories, list types, and constraints.
    
The Wikidata pattern to define a category $c$ that contains entities of type $t$ with a specific value $v$ for a property $p$ consists of a statement $(c, \msf{ category contains }, t)[p\colon v]$. Here, $p$ appears as a qualifier, although its main use is as a the main property of a statement. For example    $$\begin{array}{l}
        (\wdt{Category:Grade I listed buildings in Bedfordshire}, \\
        \ \msf{category contains}, \wdt{architectural structure}) \\
        \quad [\msf{located in the administrative territorial entity}\colon \wdt{Bedfordshire}]
        \end{array}$$
Properties that are often used to define categories include : \tf{occupation},	 \tf{performer}, \tf{sex or gender}, \tf{member of sports team},	\tf{located in the administrative territorial entity}, \tf{educated at}, \tf{place of death},  \tf{date of birth}, \tf{position held}, \tf{director, country}.
    
Similarly, a \textit{list} $l$ whose elements are of type $t$ and have a value $v$ for a property $p$ is defined by a statement of the form $(l, \msf{is a list of}, t)[p\colon v]$.
    
A \textit{constraint} on a property $p$ is defined by a set of statements of the form ($p$, \tf{property constraint}, \emph{constraint type})$[q_1: v_1, ..., q_n \colon v_n]$. For example
        $$\begin{array}{l}
             (\wdt{located on linear feature}, \msf{property constraint}, \wdt{value-type constraint})  \\
             \quad\quad [\msf{class}: \wdt{railway line},  \msf{class}: \wdt{road}, \dots]
        \end{array} $$
Typical qualifiers that appear in the definition of constraints are : \tf{class}, \tf{relation}, \tf{item of property constraint}, \tf{property scope} \tf{exception to constraint}, and \tf{constraint clarification}.

\subsubsection*{Field of a structured value or subject}

These qualifiers are used when the object of a statement is an entity that is not a Wikidata item and must be described by several fields. If one does not want to create a new item in this situation, the technique is to create a statement in which one of the fields is the object and the others are qualifiers, as in the following examples:
    \begin{itemize}
        \item  The apparent magnitude of an astronomical object $a$ is a pair $(b, f)$ where $b$ is the brightness and $f$ is the wavelength (corresponding to a filter). It is represented as 
               $ (a, \msf{apparent magnitude}, b)[\msf{astronomical filter}\colon f] $
        \item  The location of a gene in the genetic material is a number and a chromosome identifier: $(g, \msf{genomic start}, n)[\msf{chromosome}\colon  c]$
        \item  The catalog code of some entity is a string together with a  catalog identifier: ($e$, \tf{catalog code}, $s$) $[\msf{catalog}\colon c]$
        \item The boiling point of a substance is a temperature and a pressure:  $(s, \msf{boiling point}, t)$ $[\msf{under pressure}\colon p]$
    \end{itemize}

These qualifiers generally have a low to very low diversity index (< 6), i.e. they are highly specific to a property. For example, the \textsf{astronomical filter} qualifier applies almost exclusively to \textsf{apparent magnitude}. Those with a  higher index apply to several properties belonging to a common domain. For example, \tf{temperature} (diversity = 24.34) qualifies several physical values  (density, vapor pressure, solubility, etc.)

The properties they qualify generally do not make sense without the qualifiers. For instance, the position of a gene is an integer number (\tf{genomic start}) that must be qualified by the chromosome on which the gene is located.

\subsection{Additional information}
These qualifiers provide additional information, without altering the meaning of the statement.  The subcategories that we have identified are: sequence, provenance, causality, the relaiton between object/subject and statement, sub-property, external entity descriptions. 

\subsubsection*{Sequence} 

Sequence qualifiers typically indicate the absolute or relative position of the subject of a statement among the entities that are the subject of a statements with the same property and value. For example, in
              $$(\mathsf{\wdt{Jimmy\ Carter}, position\ held, \wdt{President\ of\ the\ USA}}) [\mathsf{series \ ordinal}: \wdt{39}]$$the subject (\tf{Jimmy Carter} is the 39$^{\rm{th}}$ entity that has the value \tf{President of the USA} for the property \tf{position held}. Sequence qualifiers can also apply to the value of a statement. In   $$(\mathsf{\wdt{Charlie\ Chaplin}}, \mathsf{child},  \mathsf{\wdt{Geraldine\ Chaplin}}) [\mathsf{series\ ordinal}: \wdt{4}]$$ the value (\tf{Geraldine Chaplin}) is the 4$^{\rm{th}}$ entity that is a value of the property \tf{child} for the item \tf{Charlie Chaplin}.

The most frequently used sequence qualifiers are \tf{series ordinal} (P1545),
             \tf{follows} P(155),
             \tf{followed by} (P156),
             \tf{replaces} (P1365),
             \tf{replaced by} (P1366),
             \tf{candidate number} (P4243),
             \tf{candidate position} (P10777).

\subsubsection*{Provenance} 

In Wikidata, the provenance (source) of a statement is indicated by its \emph{references}. As every statement should be based on at least one reference, Wikidata has its own specific mechanism for handling references, hence there is no reference qualifier. However, some qualifiers do provide information about the provenance of a statement. 

\tf{object named as} (P1932) and \tf{subject named as} (P1810) indicate how the statement's object and subject are name in the original source ; \tf{determination method or standard} (P459) specify the process or method or standard used to produce the value ; \tf{reason for deprecated rank} (P2241) and \tf{reason for preferred rank} (P7452) explain the ranking of a statement ; \tf{statement supported by} (P3680) provide the source of a value, as in 
$$
\begin{array}{l}
    (\tf{\wdt{Albany}, per capita income, \wdt{31,800 New Zealand dollar}})\\
    \quad [\tf{statement supported by: \wdt{2018 New Zealand census}}, \\    
    \quad \tf{point in time: \wdt{2018}}]
\end{array}$$

\subsubsection*{Causality}        
These qualifiers describe what caused the statement to become true or cease to be true, or the consequences of the statement  (i.e. the statement describes the cause of something else). The most important causality qualifiers are : \tf{end cause},
            \tf{has cause}, 
            \tf{has immediate cause}, 
            \tf{is immediate cause of}, 
            \tf{content descriptor}  (typically provides the reason/cause for a rating),
            \tf{has effect},
            \tf{cause of destruction},
            \tf{cause of death}.

\subsubsection*{Object/Subject-Statement relation}
        
These qualifiers provides a relation between the object (resp. subject) and some entity that holds in the context of the statement. This is typically the role played by the subject or object in the context of the statement, as in
$$(\wdt{2001\ A\ Space\ Odyssey},  \tf{cast\ member}, \wdt{Keir\ Dullea})[\tf{character\ role: } \wdt{David\ Bowman}]$$ 
for the object role, or in              
$$\begin{array}{l}(\wdt{Falk\ Kalamorz}, \tf{employer}, \wdt{Institute\ for\ Crop\ and\ Food\ Research})\\
\quad\quad[\mathsf{subject\ has\ role:\ } \wdt{scientist}]
\end{array}$$
for the subject role.

Typical qualifiers in this category are: \tf{object of statement has role} (P2868),
\tf{subject has role} (P2868),
\tf{character role} (P453),
\tf{position held} (P39),
\tf{position played on team / speciality} (P413).

\subsubsection*{Sub-property or sub-value (Relation or Value Refinement)}
Qualifiers such as \tf{criterion used} or \tf{mapping relation type} make the meaning of a statement more precise by providing additional information about the relation between the item and the value. For instance, in the statement   \[
\begin{array}{l}
(\wdt{\v Repka}, \msf{number of houses}, \wdt{13})\\
\quad [\msf{point in time} : \wdt{1980}, \\
\quad \msf{criterion used} : \wdt{number of permanently inhabited houses}]
\end{array}
\]
the \tf{criterion used} qualifier  indicates that \textit{number of houses} is in fact the \textit{number of permanently inhabited houses}.

Similarly, some qualifiers indicate the specific part of the value that is relevant for the statement. In \[
\begin{array}{l}
(\wdt{Promotion of science and research}, \msf{main regulatory text}, \wdt{
Fiscal Code of Germany})\\
\quad [\msf{section, verse, paragraph, or clause} :  \wdt{ \S 52 II Satz 1 Nr. 1}]
\end{array}
\]
the qualifier specifies the precise entry in the \tf{Fiscal Code of Germany} that regulates the \tf{Promotion of science and research}. 

Theses qualifiers provide a view of the statement at a finer granularity level. 

\subsubsection*{External object description}

When the object of a statement is the identifier (URL, DOI, ISBN, catalog ID, version number,  ...) of an entity that is not described in the Wikidata graph (it is not a Wikidata item), these qualifiers are used to describe this entity. In fact, they act as predicates with the external object as subject. In the following statement the \tf{language of work} qualifier describes the entity (a document) whose URL is the statement's object : \[
\begin{array}{l}
(\wdt{The Treasure of the City of Ladies}, \msf{full work available at URL}, \\
\quad\quad\quad\quad\wdt{https://www.gutenberg.org/files/26608/26608-h/26608-h.htm}) \\
\quad [\msf{language of work or name}: \wdt{French}]
\end{array}
\]
In the statement 
\[
\begin{array}{l}
(\wdt{Terraform}, \msf{software version identifier}, \wdt{1.7.2})
[\msf{publication date}: \wdt{31 January 2024}]
\end{array}
\]
the publication date is that of version 1.7.2 of the Terraform software, which is not a Wikidata item (there is only one Wikidata item that represents the software as a whole.  

Typical external object description qualifiers are : 
\tf{online access status}, \tf{author last names}, \tf{data size}, \tf{copyright license}, \tf{distribution format}.

\section{Characteristics of the taxonomy\label{sect:taxo-charact}}

\subsection{Coverage}

\begin{figure}[ht]
        \centering
        \includegraphics[width=0.99\linewidth]{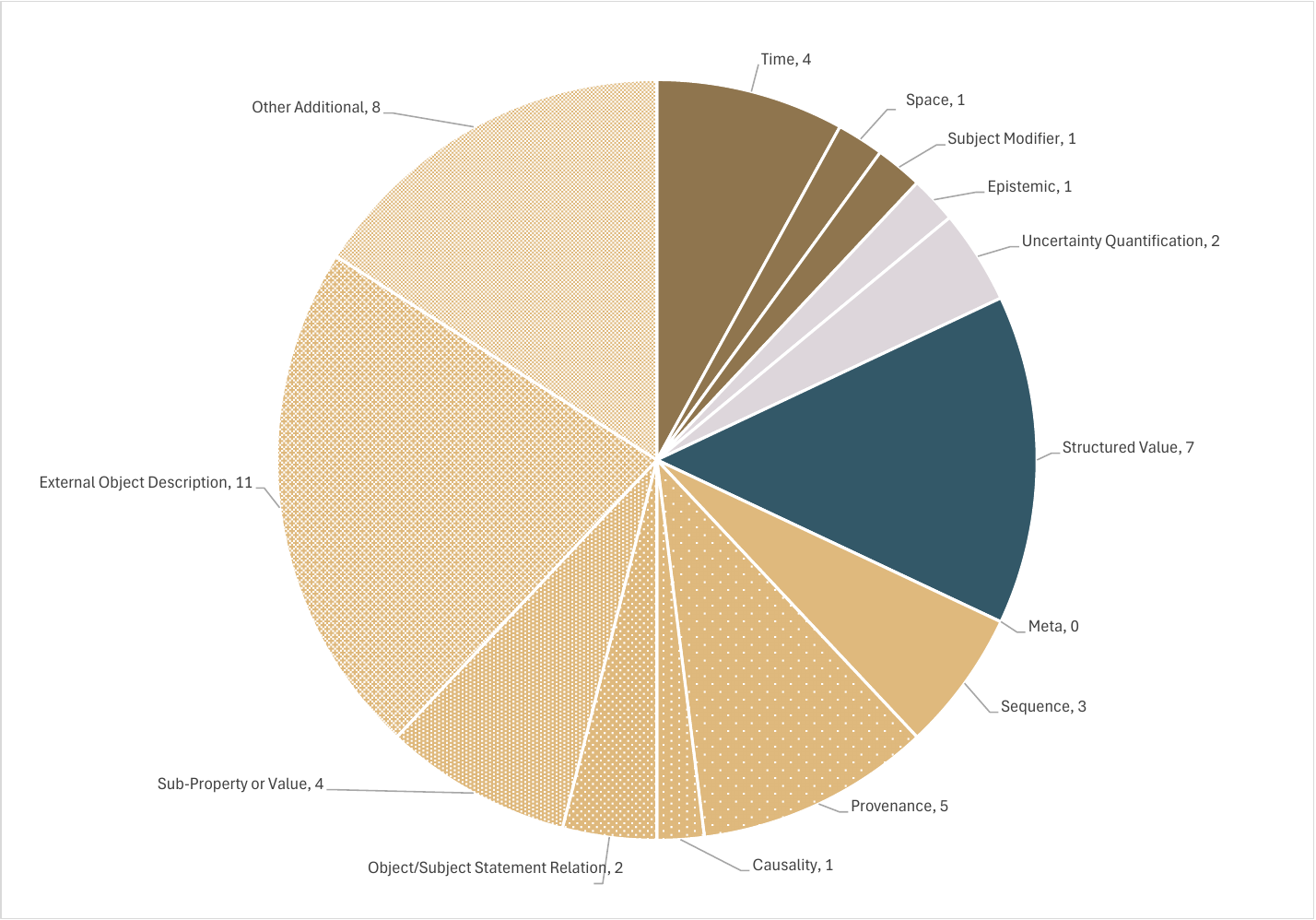}fig-
        \caption{Number of qualifiers in each category for the top 50 qualifiers}
        \label{fig:q-by-cat-50}
    \end{figure}

\begin{figure}[ht]
        \centering
         \includegraphics[width=0.99\linewidth]{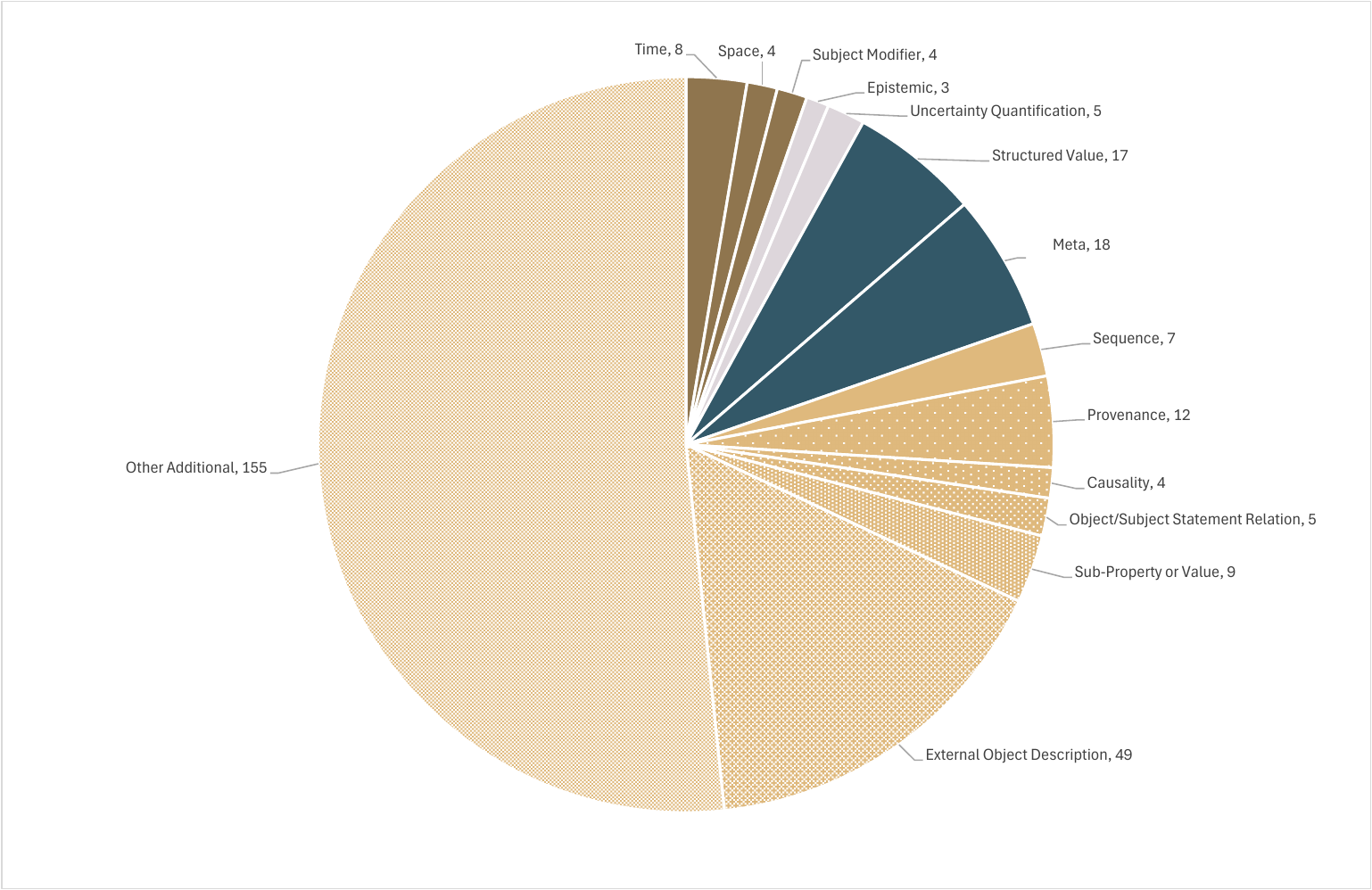}
        \caption{Number of qualifiers in each category for the top 300 qualifiers}
        \label{fig:q-by-cat-300}
    \end{figure}

Figure \ref{fig:q-by-cat-50} shows how the top 50 qualifiers\footnote{data: \path{classification/Qualifier_Analysis_2025-01-01.xlsx}} (representing 93\% of the qualifications in Wikidata) are distributed in the different taxonomy categories. There is only one category \tf{Meta} which contains no qualifiers. In the distribution of the top 300 qualifiers shown on Figure \ref{fig:q-by-cat-300}, the \tf{Other Additional} catches more than half of the qualifiers. This is because this category contains mainly domain-specific and  less important qualifiers. In fact,  qualifiers in this class have lower frequencies, as shown in Fig 2, and their average diversity is very low compared to the others, as shown in  Table 3 .

    \begin{figure}[ht]
        \centering
        \includegraphics[width=0.99\linewidth]{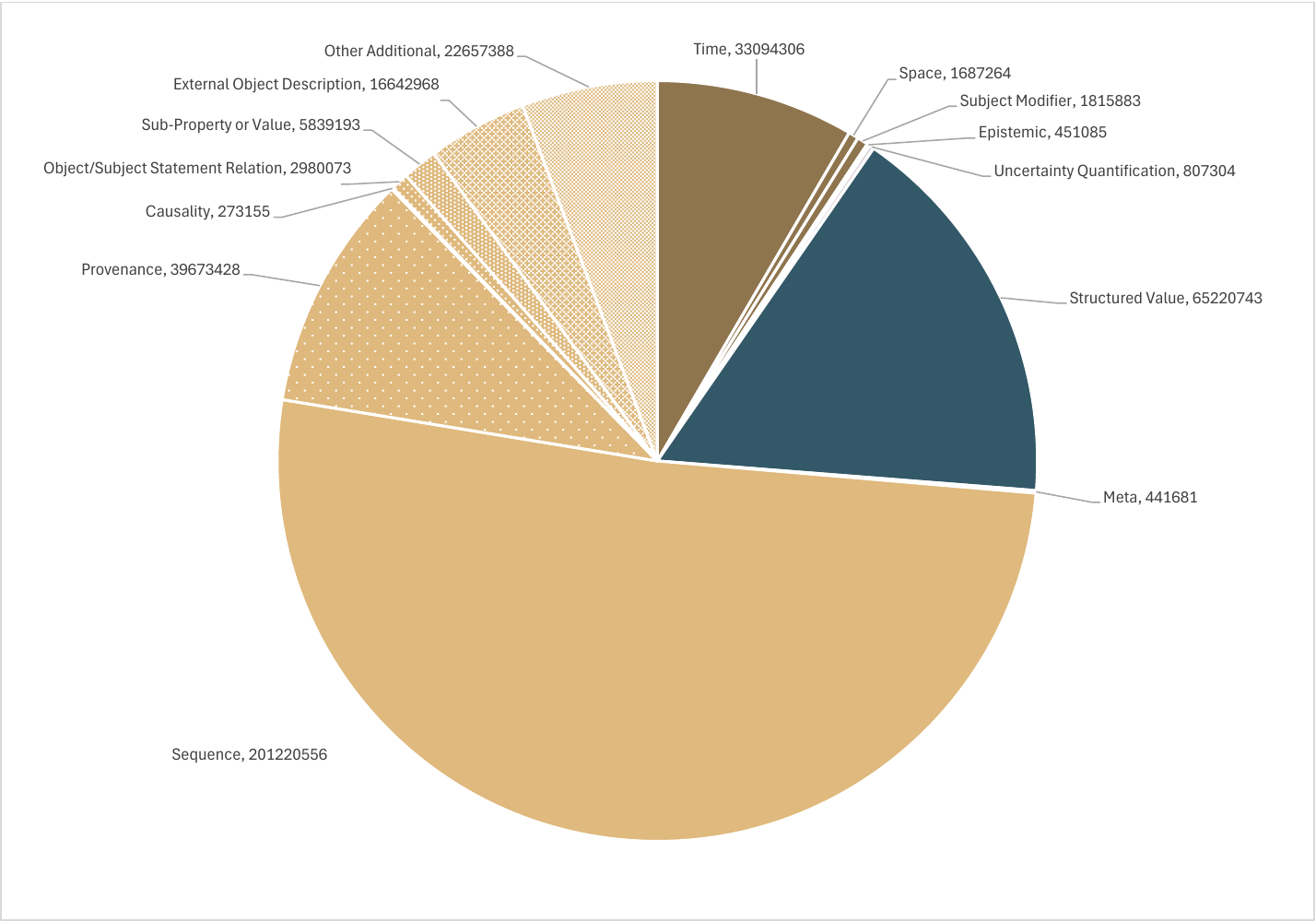}
        \caption{Sum of the frequencies of the qualifiers in each subcategory}
        \label{fig:freq-by-cat}
    \end{figure}

\begin{table}[ht]
        \centering
        \begin{tabular}{|llr|} 
        \hline     
        \textbf{Category}	&	\textbf{Subcategory} & \textbf{Average Diversity} \\
        \hline
 Context& &\\ 
 & Temporal&106.09\\ 
 & Spatial&17.90\\ 
 & Subject Modifier&65.00\\
 \hline
 Epistemic/Uncertainty& &\\ 
 & Epistemic&34.14\\
 & Uncertainty Quantification&14.36\\
 \hline
 Structural& &\\ 
 & Structured Value&7.54\\ 
 & Meta modeling&5.64\\ 
 \hline    
        Additional & & \\ 
        & Sequence&	14.82\\ 
        & Provenance&	90.40\\ 
        & Causality&	20.97\\ 
        & Object/Subject Statement Relation&	27.78\\ 
        & Sub-Property or Value&   14.49\\ 
        & External Object Description&	16.60\\ 
        
        & Other Additional&	6.13\\ 
\hline
        \end{tabular}
        \caption{Average diversity of the qualifiers for each (sub)category}
        \label{tab:cat-div}
    \end{table}

\subsection{Multiple classification / confusion}

As already mentioned, the classification of a qualifier is not always immediate and may require the observation of its actual use. However, most of the time the qualifier clearly belongs to only one class. 

For the qualifiers that fall into more than one category, we identified some possible causes

\begin{itemize}
\item 
The intended use of a property as a qualifier is not well defined or not defined at all. For instance, the \tf{country} qualifier can
     \begin{enumerate}
    \item indicate the spatial validity of a statement, as for \tf{ headquarters location} or  \tf{distributed by}or
    \item  add information about the statement's objet when used with \tf{place of birth} or \tf{location}
    \end{enumerate}
The latter case is generally a misuses. For a statement ($x$ \tf{place of birth} $y$)[\tf{country}: $c$] the country of place $y$ should normally be found in a statement ($y$ \tf{country} $c$). Therefore using a \tf{country} qualifier is redundant. Nevertheless, it occurs for \num{16191} items in Wikidata, as of 2025-01-29. This can be caused by the absence of description for \tf{country} as a qualifier and also because \tf{country} is in fact in the list of allowed qualifiers for \tf{place of birth}.
\item 
In the \tf{publisher} case, there is clearly an ambiguity in the qualifier definition. For the \tf{population} property, this qualifier indicates who published this figure, it is a provenance indication. For the \tf{described at url} property, this qualifier refers to the publisher of the web page, not to the author of the statement. It is an external entity description, not a provenance. The same occurs with  \tf{point in time}. 
\end{itemize}

\section{Usefulness of the taxonomy\label{sect:use}}

So far, we have shown that the proposed taxonomy adequately encompasses  all the selected qualifiers, meaning. each qualifier falls within a specific category. Of course the taxonomy could be refined further to identify subclasses within the Additional/Other class. However, this would mostly affect less important qualifiers. We also showed that the classes are clearly defined and non-overlapping, although in practice, some qualifiers belong to several classes due to misuse or ambiguous definitions of the qualifier). We must now return to the challenges stated in the introduction and demonstrate how this taxonomy addresses them. 

\subsection{Selecting the right qualifiers.} 

Organising qualifiers in a taxonomy helps users explore the qualifier space and find those that apply to their statements. This is true of the proposed taxonomy.
\begin{enumerate}
    \item The categories are defined by clear criteria. The definition of each category provides a way to determine whether a qualifier belongs to the category.
    \item The categories correspond to well-known notions that have been extensively studied in the field of data and knowledge representation, such as context (time, space, etc.); uncertainty; causality; provenance; ordering (sequences); and structured objects. Therefore, they are likely to provide a natural set of qualifying dimensions with which the user is familiar and which correspond to their intentions.
\end{enumerate}

\subsection{Formulating correct queries and inferences (through qualifier abstraction)} 

The abundance of qualifiers, makes it  difficult to formulate queries or inference rules that take them into account correctly and completely. The taxonomy of qualifiers can help to solve this problem by providing a higher level of abstraction.
To achieve this, each category is associated with a data type and each statement is associated with a value for each relevant category. The value for a category is formed from the set of qualifier-value pairs in which the qualifier is in that category.. For instance, the statement 
$$(s, \msf{spouse} ,v)[\msf{start time}: 1975, \msf{has cause}: marriage, \msf{end cause}: divorce, \msf{end time}: 1982]$$
has
\begin{itemize}
    \item a \textit{temporal context }value $[\msf{start time}: 1975, \msf{end time}: 1982]$
    \item a \textit{causality} value $[\msf{has cause}: marriage, \msf{end cause}: divorce]$
\end{itemize}

We can then define high level operations for these data types and use these in queries and inference rules rather than referring directly to the qualifiers.. For instance, consider an inference rule of the form 
$$(s_1, p_1, v_1)[T_1]\land(s_2,p_2,v_2)[T_2]\to(s_3,p_3,v_3)[T_3]$$
where the $T_i$'s are the temporal context values of the statement. To obtain a correct rule one can use a predicate  $\mathsf{intersects}(T_1, T_2)$, which checks if the two statements have time validity that overlap, and an operation $\msf{intersection}(T_1, T_2)$ that computes the intersection of temporal validity intervals. Thus we obtain a corrected rule
$$\begin{array}{l}
(s_1, p_1, v_1)[T_1]\land(s_2,p_2,v_2)[T_2]\land \msf{intersects}(T_1,T_2)\land \\
\to(s_3,p_3,v_3)[\msf{intersection}(T_1,T_2)]
\end{array}$$
Similarly, if the rule contains causality values $C_1$ and $C_2$, an expression $\msf{combine}(C_1, C_2)$ should be added to the rule's head to express the causality of the inferred statement. 
These operations can be quite complex, as there are multiple ways to express the same thing using qualifiers. For example, a temporal validity interval can be expressed, among other possibilities, with \tf{start time} and \tf{end time} or  \tf{start time} and \tf{duration}. Furthermore, \tf{start time} and \tf{end time} may be left unspecified, typically representing $-\infty$ and $+\infty$, respectively. However, the complexity of dealing with the qualifier values is encapsulated in the data type operation definitions and does not appear in the rules and queries.

In \cite{aljalbout2018} we propose a set of operations and their formal specification for a set of abstract qualifier types (context, sequence, .causality, provenance, additional) that are close to some of the categories and subcategories of the present taxonomy.

\subsection{Additional uses}

In addition the taxonomy can help with
\begin{itemize}
    \item Clarifying the description of each property when it is used as a qualifier. Many properties do not even have a textual description. Assigning a category to a qualifier is a good way to fix its use.
    \item Defining better user interfaces by organizing their elements according to the taxonomy (e.g. grouping the qualifiers that belong to the same category or organizing qualifiers in menus) 
\end{itemize}

\section{Related Works}

In \cite{Patel-Schneider2018} the author distinguishes between additional and contextual qualifiers in Wikidata. He shows different solutions to represent and use these qualification in RDF. In particular, he argues that contextualization qualifiers should not be shown to the users, but rather handled directly by tools such as reasoners according to a contextualization theory. By the way, our study tends to refute  the assertion found in the article that ``[...] many, perhaps most, uses of qualifiers in Wikidata and schema.org contextualize the underlying fact''. In fact, Figure \ref{fig:freq-by-cat} shows that contextualization represents less than 10\% of the qualifications.

\cite{hogan2018context,Hogan2020} survey the notion of context in knowledge graphs in terms of implementation techniques (reification, higher-arity representation, and annotation) and operations on contexts. The survey refers to \cite{zimmermann2012general} that  provides a uniform way to deal with all types of annotations by modeling annotation domains as commutative semi-rings with meet and join operations. Unlike our approach, this method does not distinguish between contextual (validity) annotations and other types of annotation; everything is treated uniformly. However, achieving this uniformity comes at the cost of a complex method for combining different domains, i.e. dealing with more than one annotation type in a query or reasoning task.

In \cite{ducu2023qualifier} the authors present an adaptation of the \tf{SchemaTree} recommender system for recommending qualifiers when creating new statements in the Wikidata knowledge base. The qualifier recommendation is not based on qualifier semantics or a pre-established taxonomy. Instead, it uses item and value type information, as well as co-occurring qualifiers.

It is also interesting to note that studies and guidelines on the design and construction of knowledge graphs generally do not address qualifier design, even in the context of property graphs. Similarly, studies of ontologies for or in Wikidata (e.g. \cite{kaffee2024ontoWd} or \cite{cente2024taxo})  are based on common ontology languages (RDFS, OWL, etc.) that do not support the notions of context, uncertainty, causality, any other type of qualification. Consequently they  do not use qualifiers.

\section{Conclusion}

This work provides an in-depth quantitative and qualitative analysis of qualifier use in Wikidata. First, we show that the importance of a qualifier is determined by two independent factors: its frequency and the diversity of its usage. We measured this diversity using a variant  of the diversity indexes employed in ecology.  Our analysis of the 300 most important qualifiers resulted in the definition of a top-level taxonomy of qualifiers that reflects how qualifiers are actually used  by the Wikidata contributors. In addition to the 'contextual' and 'additional' categories identified in previous studies we found that qualifiers can also convey epistemic information and participate in defining data structures. We also found that the previously identified categories could be refined into subcategories. 

We evaluated the taxonomy in terms of coverage, showing how the most important qualifiers are  well distributed across its categories.  In terms of usefulness, we demonstrated how this taxonomy can help to solve two problems: of 1) selecting the right qualifiers when creating new statements; and 2) formulating correct queries and inference rules on the Wikidata graph.

From a more conceptual perspective, the taxonomy highlights the main modeling dimensions  that should be considered when creating new knowledge graphs or when developing knowledge graph design methodologies.

\subsubsection*{Data Availability}
The data (frequency dictionaries), diversity values, and classification results are available on GitHub at \url{https://github.com/cui-ke/wikidata-qualifiers}.


\bibliography{bibliography}

\end{document}